\documentclass[11pt]{article}
\pdfoutput=1
\usepackage{ACL2023}

\usepackage{times}
\usepackage{latexsym}
\usepackage{diagbox}
\usepackage[T1]{fontenc}
\usepackage[utf8]{inputenc}

\usepackage{microtype}

\usepackage{inconsolata}
\usepackage{amsmath}
\usepackage{lscape}
\usepackage[utf8]{inputenc}
\usepackage{amsfonts}       % blackboard math symbols
\usepackage{nicefrac}       % compact symbols for 1/2, etc.
\usepackage{fancyhdr}       % header
\usepackage{graphicx}       % graphics\usepackage{url}
\usepackage{flushend} 
\usepackage{color, colortbl}
\usepackage{xcolor}
\definecolor{Gray}{gray}{0.9}
\newcolumntype{g}{>{\columncolor{Gray}}c}

\usepackage{diagbox} %table split headers
\usepackage{subcaption,siunitx,booktabs}
\usepackage{geometry}

\title{An Investigation of Evaluation Metrics \\for Automated Medical Note Generation}

\author{
 Asma Ben Abacha\\ 
  \texttt{Microsoft Health AI} \\
  \texttt{abenabacha@microsoft.com} \\
   \And
  Wen-wai Yim \\
  \texttt{Microsoft Health AI} \\
  \texttt{yimwenwai@microsoft.com} \\
   \AND
     George Michalopoulos \\ 
   \texttt{Microsoft Health AI} \\ 
  \texttt{georgemi@microsoft.com} \\ 
   \And
  Thomas Lin \\
  \texttt{Microsoft Health AI} \\ 
  \texttt{tlin@microsoft.com} \\
}

\begin{document}
\maketitle

%% Datasets: 
\newcommand{\mts}{\textsc{MTS-Dialog}}
\newcommand{\mediqa}{\textsc{MEDIQA-RRS}}
\newcommand{\qatriples}{\textsc{Consult-Facts}}
\newcommand{\qafull}{\textsc{Consult-Full}}

\newcommand{\hpi}{\textsc{Consult\textsubscript{HPI}}}
\newcommand{\assessment}{\textsc{Consult\textsubscript{ASSESSMENT}}}
\newcommand{\exam}{\textsc{Consult\textsubscript{EXAM}}}
\newcommand{\results}{\textsc{Consult\textsubscript{RESULTS}}}

\begin{abstract}

Recent studies on automatic note generation have shown that doctors can save significant  amounts of time when using automatic clinical note generation \cite{knoll-etal-2022-user}. Summarization models have been used for this task to generate clinical notes as summaries of doctor-patient conversations \cite{krishna-etal-2021-generating,cai-etal-2022-generation}. However, assessing which model would best serve clinicians in their daily practice is still a challenging task due to the large set of possible correct summaries, and the potential limitations of automatic evaluation metrics. In this paper,
we study evaluation methods and metrics for the automatic generation of clinical notes from medical conversations. In particular, we propose new task-specific metrics and we compare them to SOTA evaluation metrics in text summarization and generation, including: (i) knowledge-graph embedding-based metrics, (ii) customized model-based metrics, (iii) domain-adapted/fine-tuned metrics, and (iv) ensemble metrics. To study the correlation between the automatic metrics and manual judgments, we evaluate automatic notes/summaries by comparing the system and reference facts and computing the factual correctness, and the hallucination and omission rates for critical medical facts. This study relied on seven datasets manually annotated by domain experts. Our experiments show that automatic evaluation metrics can have substantially different behaviors on different types of clinical notes datasets. However, the results highlight one stable subset of metrics as the most correlated with human judgments with a relevant aggregation of different evaluation criteria. 

\end{abstract}

\section{Introduction}

In recent years, the volume of data created in healthcare has grown considerably as a result of  record keeping policies \cite{Kudybabook}. The documentation requirements for electronic health records  significantly contribute to physician burnout and work-life imbalance \cite{Arndt419}. Automatic generation of clinical notes can help healthcare providers by significantly reducing the time they spend on documentation, and allowing them to spend more time with patients \cite{PAYNE201891}. It can also improve the clinical notes' accuracy by reducing errors and inconsistencies in documentation, leading to patient records with higher quality.

A reliable evaluation methodology is necessary to build and improve clinical note generation systems, but faces the two traditional limitations of evaluating Natural Language Generation (NLG) systems. On one hand, human-expert evaluation, considered to be the most reliable way to evaluate NLG systems, can be both time-consuming and expensive. On the other hand, evaluating the performance of natural language generation (NLG) systems automatically can be challenging due to the complexity of human language. 

Several metrics have been proposed to evaluate the performance of NLG systems, including lexical N-gram based metrics and embedding-based metrics that measure the similarity between a system's generated text and one or more reference texts using pre-trained language models.

While several research efforts studied and compared automatic evaluation metrics on many open-domain and domain-specific datasets such as the CNN/DailyMail and TAC datasets \cite{Lin04-eval,owczarzak-etal-2012-assessment,peyrard-2019-studying,FabbriKMXSR21,naacl-DeutschDR22}, very few research works addressed the adequacy of evaluation metrics to the task of clinical note generation, where e.g., omitting critical medical facts in the generated text is a more significant failure point. To the best of our knowledge, only one research paper addressed this task based on one synthetic dataset of 57 mock consultation transcripts and summary notes \cite{ACL-Moramarco22-Eval}.

In this paper, we study evaluation methods and metrics for the automatic generation of clinical notes from medical conversations, including their correlations with human assessments of factual omissions and hallucinations. We also propose new task-specific metrics and we compare them to SOTA evaluation metrics across several clinical text summarization datasets. 

\noindent Our contributions are as follows:
\begin{itemize}
\item We study the relevance and impact of a wide-range of existing automatic evaluation metrics in clinical note generation.
\item We propose and study four types of evaluation metrics for the task of automatic note generation: knowledge-graph embedding-based metrics, customized model-based metrics, domain-adapted/fine-tuned metrics, and ensemble metrics\footnote{We publish the source code and fine-tuned checkpoint at: \url{https://github.com/abachaa/EvaluationMetrics-ACL23}}. 
\item We compare these metrics with SOTA metrics by performing a wide evaluation with 21 metrics according to different criteria such as factual correctness, hallucination, and omission rates.
\item To perform a fact-based evaluation of the generated notes, we annotate seven datasets of automatically generated clinical notes using key phrase- and fact-based annotation guidelines that we use to compute reference manual scores for the correlation study\footnote{We also release the manual annotations: \url{https://github.com/abachaa/EvaluationMetrics-ACL23}}.
\end{itemize}

%===========================================
\section{Related Work}
%===========================================

Different evaluation metrics are commonly used to evaluate text summarization and generation including ROUGE-N \cite{lin-2004-rouge}, BERTScore \cite{bertscore}, MoverScore \cite{moverscore}, BARTScore \cite{Bartscore}, and BLEURT \cite{bleurt}. Other metrics have been also proposed for evaluating factual consistency and faithfulness  \cite{FEQA, MaynezNBM20, WangCL20, FRANK, abs-2209-03549}. 

To study their effectiveness, several efforts focused on comparing automatic metrics such as ROUGE and BLEU  based on their correlation with human judgments \cite{graham-2015-evaluating}, and showed that automatic evaluation of generated summaries still has several limitations and biases \cite{Hardy2019, Re-evaluatingSumEval2021}. 
Furthermore, in \cite{bhandari-etal-2020-evaluating}, the authors  showcase that  the effectiveness of an evaluation metric depends on the task (e.g. summarization) and on the application scenario (e.g. system-level/ summary level).

Despite observations of frequent disagreements in manual evaluation campaigns \cite{NLG_20Years_2020}, expert-based evaluation remains an effective method to assess the performance of automatic metrics, especially in specialized domains. However, it relies on the availability of domain experts to rate the summaries and relevant datasets. Recently, \cite{ACL-Moramarco22-Eval} studied the task of medical note generation on a small set of 57 transcript-note pairs, manually annotated by clinicians. Their experiments showed that character-based Levenshtein distance, BERTScore, and METEOR performed best for evaluating automatic note generation in that dataset.

%===========================================
\section{Evaluation Methodology}
%===========================================
%%%======
\begin{table*}[h]
\centering
\footnotesize
\begin{tabular}{lccccc}
\hline
Dataset & \#Summary & \#Words & \#Words  & Annotations\\
& Pairs & /Summary & /Reference  & \\
\hline
\mts{} & 400 &  15 &36 &   Facts  \\
\mediqa{} & 182 & 18  & 28& Facts  \\
\qatriples{} & 54 & 203 &214& Facts \\
\hpi{} & 3,397  & 333& 336& Key Phrases \\
\assessment{} &  3,141 & 149 & 177 & Key Phrases\\
\exam{} & 2,144 & 163 & 137& Key Phrases\\
\results{} & 540 &  38 &15 & Key Phrases\\
\hline
\end{tabular}
\caption{Annotated datasets of clinical summaries and reference notes.}
\label{tab:datasets}
\end{table*}

%%%======
To assess the relevance and suitability of automatic evaluation metrics for the task of clinical note generation, we create expert-based annotations for critical aspects such as factual consistency, hallucinations, and omissions. We then assess each metric in light of its correlation with manual scores generated from the expert annotations.

%=================================================
\subsection{Fact-based Annotation}
%================================================

We define a fact as information that cannot be written in more than one sentence (e.g., \textit{"Family history is significant for coronary artery disease.}"). Medical facts include problems, allergies, medical history, treatments, medications, tests, laboratory/radiology results, and diagnoses. We also include the patient age, gender, and race, and expand the critical facts to the patient and his family. 

Annotators extracted individual facts from both the reference and system summaries in the form of subject-predicate-object expressions and following the above fact definition. 
 
 Comparing between a reference and hypothesis summary, and referencing the source text/conversation if required, the annotators were additionally tasked to identify overlapping and non-overlapping facts to one of several categories which were later automatically counted. These included:
 \begin{itemize}
     \item Critical Omissions: the number of medical facts that were omitted, 
     \item  Hallucinations: the number of hallucinated facts.  Hallucinations are factual errors that do not exist in the source text and cannot be supported by the source facts (e.g., added dates, names, or treatments). 
     \item Correct Facts: the number of correct facts according to the input conversation and the reference summary, and 
     \item Incorrect Facts: the number of incorrect facts outside of hallucinations. Incorrect facts include values and attributes that are incorrectly copied from the source (e.g., date with a wrong year, wrong age, or dose). 
 \end{itemize}

Three trained annotators with medical background participated in the annotation process. Inter-annotator agreement for these computations are shown in Table~\ref{tab:iaa_qa} and Table~\ref{tab:iaa_rrs} in Appendix A.

%=================================================
\subsection{Key Phrase-based Annotation}
%================================================

The key phrase- and fact-based annotations use different ways of representing information in clinical notes. While the fact-based annotation compares semantic triples (e.g., "Back pain stopped 8 days ago" vs. "Low back pain started 8 years ago"), the key-phrase annotations involved labeling incorrect words and phrases; for instance: "back pain" (instead of "lower back pain"), "stopped" (instead of "started"), or "8 days" (instead of "8 years"). This method is more conducive in a production environment where errors can be attributable to specific parts of the report; the same labeling method is often also used for feedback to the author of the note in our different human quality review settings. 

In our annotation setups, the key phrase-based annotation operated on text span highlights, while the fact-based annotation required more steps as the annotators were required to write the system and reference facts based on the system and reference summaries before comparing their counts. 

Using highlights, critical hallucinations and incorrect information can be identified; meanwhile omissions were marked by identifying a required insertion of information in a corresponding location of the note. However, unlike the previous annotation, repeats of the same incorrect facts may be counted more than once if they appear multiple times. The labels produced here were from \qafull{} dataset (cf. Section~\ref{sec:datasets}), with a reported average agreement of critical hallucinations, omissions, and inaccuracies was at 0.80 F1 score, relaxed overlap between 12 annotator pairs.

%% - got the average from 018 first set - https://nuancewiki.atlassian.net/wiki/spaces/ACWM/pages/1639229/QA+Analysis+Q3+FY22#QAAnalysisQ3FY22-018-Firstsetof10encounters

\subsection{Reference Scores}

From the fact-based annotations, we compute the following reference scores: 

\[FactualPrecision=\frac{\#CorrectFacts}{\#SystemFacts}\]
%\vspace{-8mm}
\[Factual~Recall = \frac{\# Correct~Facts}{\# Reference~Facts}\]
\[Hallucination~Rate = \frac{\# Hallucinated~Facts}{\# System Facts}\]
\[Omission~Rate = \frac{\# Omitted~Facts}{\# Reference~Facts}\]

%\[\mbox{\footnotesize\displaystyle\[Omission~Rate = \frac{\# Omitted~Facts}{\# Reference~Facts}\]}\]
\begin{itemize}
\item \it{\footnotesize{System Facts = Correct + Incorrect + Hallucinated}}  
\end{itemize}

\noindent From the key phrase-based annotations, we compute the normalized hallucination and omission counts:
\[Hallucination Count = \frac{\# Hallucinated~key~phrases}{\# System~Summary~Words}\]
\[Omission Count = \frac{\# Omitted~key~phrases}{\# Reference~Summary~Words}\]

\subsection{Datasets}
\label{sec:datasets}

Publicly available datasets on medical note generation and clinical text summarization are rare compared to open-domain data. For this study, we use three main  collections: 
\begin{itemize}
\item The \mts{} collection of 1.7k pairs of doctor-patient dialogues and associated clinical notes \cite{mts-dialog}. System summaries are generated using the BART model \cite{BART}. 
\item The \mediqa{} dataset includes 182 pairs of clinical notes and system summaries randomly selected from the MEDIQA-RRS collection \cite{mediqa2021}. 
\item An in-house collection of medical notes (called \qafull{}) from multiple specialties with system summaries generated using a pointer-generator transformer model from doctor-patient conversations \cite{enarvi-etal-2020-generating}. 
\end{itemize}

We followed the fact-based annotation guidelines to annotate the \mts{} and \mediqa{} datasets, and a random subset from the \qafull{} collection, called \qatriples{}. %% QA-Review-Triples 

To study the relevance of the automatic metrics to the individual sections of clinical notes, we also split the \qafull{} collection into four subsets: \hpi{}, \assessment{}, \exam{}, and \results{}, which include summaries associated with the HPI, Assessment, Exam, and Results sections, and we annotated them manually at a phrase level.  %% QA-Review-Dataset 

Table~\ref{tab:datasets} provides statistics about the datasets.  

%===========================================
\section{Task-specific Evaluation Metrics}
\label{sec:metrics}
%===========================================

We study four different types of evaluation metrics for the task of automatic clinical note generation, that take into account the specificities of the medical domain by:  (i) using embeddings built from medical Knowledge graphs (e.g, UMLS), (ii) adapting model-based metrics (e.g., BERTScore) by increasing the weights of medical terms, (iii) fine-tuning a model-based metric on a large collection of clinical notes, and (iv) building linear ensembles based on normalization and averaging of different metrics.

%=================================================
\subsection{Knowledge-Graph Embedding-based Metrics}
%=================================================

Our first approach, called MIST, relies on knowledge embeddings generated by a Knowledge-Graph Embedding (KGE)-based model. Knowledge graphs provide additional semantic information that can support language understanding, especially in the medical domain where both terminologies and facts might not be common enough to be captured by contextual embeddings.  

To build medical KGE, we use a generative adversarial networks model \cite{naacl-KBGAN-18} trained on concepts and relations from the Unified Medical Language System (UMLS) \cite{umls-Lindberg93,umls-Bodenreider04}. 

The MIST metric relies on the embeddings of the medical concepts recognized in the texts to compute the similarity between the reference clinical notes and the automatically generated summaries.  

To link the clinical notes to the UMLS concepts, we extract medical concepts by combining the scispaCy \cite{neumann-etal-2019-scispacy} and MedCAT \cite{Kraljevic2021-ln} entity linking models. 

We compute the final recall-oriented MIST value using the graph-based embeddings ($G_c$) of each concept $c$ recognized in the reference and system summaries and the cosine similarity, as follows, for a set of reference concepts $R$ and a set of system concepts $S$:
\begin{equation}
MIST(S,R) = \frac{1}{|R|} \sum_{c \in S} max_{r \in R} ~~cos(G_c, G_r)
\end{equation}
\vspace{-8mm}

%================================================
\subsection{Finetuning-based Metric}
%================================================
%% slide notes are here: https://nuancecommunications-my.sharepoint.com/:p:/r/personal/msft_wen-wai_yim_nuance_com/_layouts/15/Doc.aspx?sourcedoc=%7B385E3DAE-DDD2-490C-B0A2-485E5C09ECF4%7D&file=bleurt_qafintetuning.pptx&action=edit&mobileredirect=true

Our second approach relies on fine-tuning model-based metrics on relevant large medical collections of family medicine and orthopaedic notes. In particular, we started with the BLEURT-512 model \cite{bleurt} and fine-tuned it using a quality score, derived from an assigned \textit{error score}\footnote{This error score is calculated by a weighted sum of critical and non-critical errors, as well as spelling/grammar/style errors annotated by domain expert labelers. The weight scheme is given in Appendix B, Table \ref{tab:errorweights}.} from an internal quality review grading. The derived \textit{quality score} was calculated by the following equation:
\begin{equation}
\small
quality = 1 - \frac{error\_score}{max\_sentlen(summary,reference)}
\end{equation}
\indent A total of 6,367 family medicine and orthopaedic encounters were used for fine-tuning. To maximize diverse pairings as well as to satisfy BLEURT's maximum sequence length constraint, we fine-tuned at the level of each note's HPI, EXAM, RESULTS, and ASSESSMENT sections (with empty sections removed), resulting in 17,852 pairs. We fine-tuned over one epoch at default parameters. We call the resulting metric based on this model: ClinicalBLEURT.

%=================================================
\subsection{Customized Model-based Metrics}
%================================================

\subsubsection{Medical Weighted Evaluation Metrics}
Our third approach relies on designing new customized model-based metrics that assign 
a higher weight to the term with a medical meaning. These medical weighted metrics will allow us to examine whether words with a medical meaning  can be more indicative for sentence similarity than common words for the task of  automated medical note generation. Specifically, we update the scoring policy of two popular  evaluation metrics, by providing a higher weight to the words in the summaries that have a  medical meaning: \\(i) BARTScore \cite{Bartscore} which uses a  seq-seq model to calculate 
the log probability of one text $y$   given another text $x$, and \\(ii) BERTScore \cite{bertscore} which  computes a similarity score for each token in the candidate summary with each token in the reference. \\
For both metrics, firstly we identify  all the words, in the candidate and in the reference summary, which have a clinical meaning defined in UMLS using the MedCAT toolkit \cite{Kraljevic2021-ln}. We then modify the scoring policy of both evaluation metrics to a weighted scoring policy  where the weight for all the medical words is higher to provide a stronger incentive to the evaluation model to take in consideration these words during the evaluation of a candidate summary. Specifically, the BARTScore metric  is updated to:
\vspace{-3mm}
\begin{equation}
MedBARTScore = \sum_{i=1}^m w_t \log p(y_t|y<t, x ) 
\label{eq:bartscore}
\end{equation}
\vspace{-8mm}
% BART_{weight} = \sum_{i=1}^m w_t \log p(y_t|y<t, x ) 

where $x$ is the source sequence and $y=(y_1,...,y_m)$ are the tokens of  the target sequence of length $m$. 

We also update the BERTScore for a pair of a reference summary $x$ and candidate summary $\hat{x}$  to: 
\vspace{-3mm}
\begin{align*} 
MedBERTScoreP  = \frac{1}{ \hat{|x|}} \sum_{ \hat{x_i} \in  \hat{x}} w_x \max_{x_j} x_i^\top \hat{x_j} \\
\label{eq:bertscore}
\end{align*} 
where, for both metrics, $w=1$ for all the non-medical words and $w_t=1 +\alpha$ for all the   words with a medical meaning, where  $\alpha$ is an additional weight value for these words. After experimenting with different values in the $[0.1, 1.5]$ range, we found that the best $\alpha$ value was $1.0$ for the weight policy.

%%% BERT_{weight}R = \frac{1}{|x|} \sum_{x_i \in x} w_x \max_{x_j} x_i^\top \hat{x_j}  \\
%% BERT_{weight}F = 2 \frac{BERT_{w}P * BERT_{w}R}{BERT_{w}P + BERT_{w}R} 
\subsubsection{Sliding Window Policy}
The main disadvantage of the previously mentioned model-based metrics over the traditional evaluation metrics (e.g., ROUGE) is that they can only encode texts that have length less than the encode-limit of the pre-trained models that are based on.  For example, the encode-limit for a BERT-based metric is 512 tokens. However, real-world summaries and clinical notes may contain more than 512 tokens. For instance, our analysis in the \qafull{} dataset shows that  $31\%$ of the summaries have more than 512 tokens. 
We, therefore, create a variation of the BERTScore metric where we use a sliding window approach with the offset size of 100 tokens to encode overlength summaries. 

Our sliding window policy is to first split the initial sentence into segments of at most 512 tokens with an overlap size of 100 tokens. Afterward, we calculate the embeddings of these segments independently and concatenate the results to get the original document representation. 

This metric will be referred to as MedBERTScore-SP in the Results section. 

%% For the calculation of the Bertscore, MedBertScore-P, and MedBertScore-SP, we use the embeddings created by the last layer of the contextual model. 

%=================================================
\subsection{Ensemble Metrics}
%================================================

To take advantage of the different perspectives brought by knowledge graph-based metrics, contextual embedding-based metrics, and lexical metrics, we tested different ensembles of normalized metric values. We selected the top-2 performing ensemble metrics for further experiments; $MIST_{Comb1}$ and $MIST_{Comb2}$:  
\vspace{-3mm}
\begin{equation} 
Z_m(x) = \frac{x - \mu_m}{\sigma_m}
\end{equation}
\begin{equation} 
MIST_{Comb1}(x) = \frac{1}{3}\sum_{m \in C_1} Z_m( m(x) ) 
\end{equation}
\begin{equation} 
MIST_{Comb2}(x) = \frac{1}{3}\sum_{m \in C_2} Z_m( m(x) ) 
\end{equation}

with $Z_m(x)$ the normalized $Z_{score}$ of a metric $m$, $\mu_m$ the mean value of $m$ over the summaries set, $\sigma_m$ the standard deviation of $m$, $C_1 = \{$MIST, ROUGE-1-R, BERTScore$\}$ and $C_2 = \{$MIST, ROUGE-1-R, BLEURT$\}$

%=================================================
\section{Evaluation Setup}
%=================================================

We used the deberta-xlarge-mnli  model \cite{he2021deberta} as the base model for BERTScore and the BLEURT-20 checkpoint for the BLEURT metric, that correlate better with human judgment than the default variants based on recent experiments. For BARTScore metric, we used the BART model  that was trained on the  ParaBank2 dataset \cite{hu-etal-2019-large} which was provided by the authors.

From the designed and tested 50+ metrics and variants (e.g. our new metrics and variants, open-domain metrics, ensemble metrics), we selected the top 21 metrics to study and analyze their performance on the different datasets. The selection was based on the performance of these metrics and their Pearson correlation scores with human judgments on the \mts{} and the \qafull{} datasets. 
Our first tests also included open-domain fact-based metrics such as FactCC \cite{kryscinskiFactCC2019} (trained on the CNN/DailyMail dataset) and QA metrics such as QUALS \cite{QUALS} (developed using XSUM and CNN/DailyMail) but they did not perform well due to the differences between open-domain and clinical questions/answers. %Adapting these metrics to the clinical domain is expensive and requires creating and annotating large training datasets of clinical notes.

The experiments were performed on one 80GB NVidia A100 GPU. 

%=================================================
\begin{table*}[] %% DONE! 
\centering
\footnotesize 
 \scalebox{0.9}{
\begin{tabular}{lccc|cc|c}  
\hline 
\diagbox[dir=NW]{\rule{0mm}{0.03cm}\rule{0.03cm}{0cm}Automatic}{Reference} & $\uparrow$ \textbf{Factual P}  & $\uparrow$ \textbf{Factual R}  & $\uparrow$ \textbf{Factual F1}  & $\downarrow$ \textbf{Hallucination}  & $\downarrow$ \textbf{Omission} & $\uparrow$  \textbf{Aggregate Score}\\ 
\hline  
\textbf{SOTA Metrics} &  &  &  &  & \\ 
ROUGE-1-P & 0.14 & -0.09 & -0.04 & -0.16 & 0.06 & 0.00 \\
ROUGE-1-R & 0.10 & 0.57 & 0.53 & 0.02 & -0.60 & 0.41\\
ROUGE-1-F & 0.13 & 0.39 & 0.40 & -0.08 & -0.44 & 0.33 \\
ROUGE-2-P & 0.12 & 0.05 & 0.07 & -0.12 & -0.12 & 0.10\\
ROUGE-2-R & 0.12 & 0.34 & 0.34 & -0.09 & -0.39 & 0.29 \\
ROUGE-2-F  & 0.12 & 0.28 & 0.29 & -0.10 & -0.33 & 0.25\\ 
ROUGE-L-P & 0.13 & -0.08 & -0.05 & -0.15 & 0.07 & 0.00 \\
ROUGE-L-R & 0.10 &  0.56 & 0.51 & 0.02 & -0.58 & 0.40\\
ROUGE-L-F  & 0.13 & 0.38 & 0.38 & -0.08 & -0.41 & 0.31 \\
BERTScore-P & 0.10 & 0.11 & 0.15 & -0.18 & -0.23 & 0.18\\ 
BERTScore-R & 0.07 & \underline{0.62} & \underline{0.59} & 0.02 & \bf -0.71 & \bf 0.47\\ 
BERTScore-F & 0.09 & 0.44 & 0.45 & -0.08 & -0.56 & 0.38\\
BLEURT & 0.11 & 0.48 &  0.47 & -0.08 & -0.59 & 0.40 \\ 
BARTScore & \underline{0.37} & 0.09 & 0.19 & \underline{-0.34} & -0.26 & 0.25\\
\hline 
\textbf{New Metrics} &  &  &  &  & \\ 
MedBERTScore-P &  0.28 & -0.16 & -0.02 & -0.27 & -0.32 & 0.14\\ 
MedBERTScore-SP &  0.28 & -0.16 & -0.02 & -0.27 & -0.32 & 0.14\\
MedBARTScore & \bf 0.46 & 0.13 & 0.24 & \bf -0.46 & -0.27 & 0.30\\ 
ClinicalBLEURT & 0.19 & 0.22 &  0.19 & -0.06 & -0.20 & 0.16\\ 
MIST &  0.02 &  0.46 & 0.45 &  0.08 & -0.51 & 0.33\\
MIST-Comb1 & 0.08 & \bf 0.64 & \bf 0.61 & 0.05 & \bf -0.71 & \bf 0.47\\ 
MIST-Comb2 & 0.09 & 0.60 & 0.58  &  0.01 & -0.68 & 0.46 \\ 
\hline  
\end{tabular}
}
\caption{\textbf{\mts{}}: Pearson's correlation coefficients between the automatic and manual scores. Best results are highlighted in bold and second best are underlined.}
\label{tab:corr_mts}
\end{table*}
%===========================
\begin{table*}[] %% DONE! 
\centering
\footnotesize 
\scalebox{0.9}{
\begin{tabular}{lccc|cc|c}  
\hline 
\diagbox[dir=NW]{\rule{0mm}{0.03cm}\rule{0.03cm}{0cm}Automatic}{Reference} & $\uparrow$ \textbf{Factual P}  & $\uparrow$ \textbf{Factual R}  & $\uparrow$ \textbf{Factual F1}  & $\downarrow$ \textbf{Hallucination}  & $\downarrow$ \textbf{Omission} & $\uparrow$  \textbf{Aggregate Score}\\ 
 \hline
\textbf{SOTA Metrics} &  &  &  &  & \\ 
ROUGE-1-P & 0.63 & 0.32 & 0.50 & \bf{-0.73} & -0.46 & 0.55\\
ROUGE-1-R & 0.59 & \bf{0.80} & \bf{0.79} & -0.39 & -0.84 & 0.70 \\
ROUGE-1-F & \bf 0.70 &  0.70 & 0.78 & -0.55 & -0.79 & \bf 0.73\\ 
ROUGE-2-P & 0.56 & 0.33 & 0.45 & -0.60 & -0.43 & 0.48 \\
ROUGE-2-R & 0.55 & 0.73 & 0.71 &-0.39 &  -0.78 & 0.65\\
ROUGE-2-F  & 0.62 & 0.62 & 0.68 & -0.49 & -0.70 & 0.64\\ 
ROUGE-L-P & 0.63 & 0.33 & 0.51 & \bf -0.73 &  -0.47 & 0.56\\
ROUGE-L-R & 0.60 & \bf{0.80} & \bf{0.79} & -0.40 & -0.84 & 0.71\\
ROUGE-L-F  &  \bf 0.70 & 0.70 & 0.78 & -0.56 & -0.79 & \bf 0.73\\ 
BERTScore-P & 0.62 & 0.47 & 0.58 & -0.56 & -0.60 & 0.58\\ 
BERTScore-R & 0.60 & \bf{0.80}  & 0.78 & -0.37 & \bf{-0.85} & 0.70\\ 
BERTScore-F & 0.66 & 0.69 & 0.74 & -0.49 & -0.79 & 0.69\\ 
BLEURT  & 0.61  & 0.67 & 0.71 & -0.49 & -0.76 & 0.67\\ 
BARTScore & 0.61 & 0.34 & 0.51 & -0.66 & -0.41 & 0.52\\ 
\hline 
\textbf{New Metrics} &  &  &  &  & \\ 
MedBERTScore-P & 0.63 & 0.47 & 0.59 & -0.57 & -0.60 & 0.59\\ 
MedBERTScore-SP & 0.63 & 0.47 & 0.59 & -0.57 & -0.61 & 0.59\\ 
MedBARTScore & 0.61 & 0.35 & 0.51 & -0.67 & -0.42 & 0.53\\ 
ClinicalBLEURT & 0.04 & 0.15 & 0.08 & 0.09 & -0.15 & 0.05\\ 
MIST & 0.08 & 0.44 & 0.31 & 0.08 & -0.44 & 0.25\\
MIST-Comb1 & 0.48 & 0.78 & 0.72 & -0.26 & -0.81 & 0.63\\ 
MIST-Comb2 & 0.53 & \bf{0.80} & 0.75 & -0.33 & \bf{-0.85} & 0.67\\ 
\hline  
\end{tabular}
}
\caption{\textbf{\qatriples{}}: Pearson's correlation coefficients between the automatic and manual scores. }
\label{tab:corr_qatriples} 
\end{table*}

%============
\begin{table*}[] %% DONE! 
\centering
\footnotesize 
\scalebox{0.95}{
\begin{tabular}{lccc|cc|c}  
\hline 
\diagbox[dir=NW]{\rule{0mm}{0.03cm}\rule{0.03cm}{0cm}Automatic}{Reference} & $\uparrow$ \textbf{Factual P}  & $\uparrow$ \textbf{Factual R}  & $\uparrow$ \textbf{Factual F1}  & $\downarrow$ \textbf{Hallucination}  & $\downarrow$ \textbf{Omission} & $\uparrow$  \textbf{Aggregate Score}\\ 
\hline  
\textbf{SOTA Metrics} &  &  &  &  & \\
ROUGE-1-P & \bf 0.40 & -0.10 & 0.00	& \bf -0.39	& -0.30 & 0.17 \\	
ROUGE-1-R & 0.22 &	0.55 & 0.57	& -0.22	& \bf -0.74 & 0.53\\
ROUGE-1-F & 0.31 & 0.39	& 0.47	& -0.31	& -0.69 & 0.49 \\ 
ROUGE-2-P & 0.34 &	0.04	& 0.10	& -0.32	& -0.36 & 0.22 \\
ROUGE-2-R & 0.20 &	0.46 & 0.47 & -0.18 & -0.66 & 0.45\\ 
ROUGE-2-F  & 0.25	& 0.37	& 0.41	& -0.23	& -0.63 & 0.42\\ 
ROUGE-L-P &  0.37	& -0.11	& -0.02 &	-0.36	& -0.29 & 0.15\\
ROUGE-L-R & 0.20 &  0.54 &  0.55  &	-0.21	& \underline{-0.73} & 0.51\\ 
ROUGE-L-F  & 0.29	& 0.38 &	0.44 &	-0.29	& -0.69 & 0.47\\  
BERTScore-P & 0.31	& -0.07	& 0.03	& -0.30	& -0.33 & 0.17\\ 
BERTScore-R & 0.17	& 0.56 & 0.58 & -0.21	& \underline{-0.73} & 0.53\\  
BERTScore-F & 0.29	& 0.32	& 0.38	& -0.30	& -0.64 & 0.43\\  
BLEURT  & 0.33	& 0.46 &	0.51	& -0.29	& -0.69 & 0.50 \\ 
BARTScore & \underline{0.38}	& 0.15	& 0.23	& \underline{-0.37}	& -0.39 & 0.31 \\ 
\hline 
\textbf{New Metrics} &  &  &  &  & \\ 
MedBERTScore-P &  0.32	& -0.04	& 0.05	& -0.31 &	-0.35 & 0.19\\
MedBERTScore-SP &  0.32	& -0.04	& 0.05	& -0.31 &	-0.35 & 0.19\\
MedBARTScore & 0.29	& 0.03	& 0.13	& -0.28 &	-0.30 & 0.21\\
ClinicalBLEURT &  0.27 & 0.11 & 0.10 & -0.26 & -0.09 & 0.14\\ 
MIST & 0.11 &	\bf 0.73	& \bf 0.66	& -0.10	& -0.52 & 0.49 \\
MIST-Comb1 &  0.18	& \underline{0.67} &	\bf 0.66 &	-0.19	& -0.72 & \bf 0.56\\ 
MIST-Comb2 & 0.24	& 0.64	& 0.65	& -0.23 &	-0.72 & \bf 0.56\\
\hline  
\end{tabular}  
}
\caption{\textbf{\mediqa{}}: Pearson's correlation coefficients between the automatic and manual scores. Best results are highlighted in bold and second best are underlined.}  
\label{tab:corr_mediqarrs}
\end{table*}

%%%======
\begin{table*}[]
\centering
\footnotesize 
\scalebox{0.9}{
\begin{tabular}{lcc|cc|cc|cc}  \hline 
& \multicolumn{2}{c}{HPI Section}  & \multicolumn{2}{c}{Assessment Section} & \multicolumn{2}{c}{Exam Section}  & \multicolumn{2}{c}{Results Section}   \\\hline 
& Hallucination & Omission & Hallucination & Omission & Hallucination & Omission & Hallucination & Omission \\ 
\hline 
\textbf{SOTA Metrics} &  &  &  &  &  &  &  &  \\ 
ROUGE-1-P & -0.23 & -0.21 &  -0.45 &  \bf -0.30 & -0.19 & -0.17 & -0.18 & -0.23 \\ 
ROUGE-1-R & -0.20 & -0.18 & -0.33 & -0.21 & -0.19 & -0.15 & -0.09 & -0.19 \\ 
ROUGE-1-F & -0.24 & -0.22 & -0.37 & -0.25 & -0.21 & -0.18 & -0.11 & -0.20 \\ 
ROUGE-2-P & -0.25 & -0.21 & -0.46 & \bf -0.30 & -0.24 & -0.17 & -0.18 & \bf -0.24 \\
ROUGE-2-R & -0.22 & -0.19 & -0.37 & -0.24 & -0.21 & -0.18 & -0.12 & -0.23 \\ 
ROUGE-2-F  & -0.25 & -0.21 & -0.41 & -0.27 & -0.23 & -0.18 & -0.13 & -0.22 \\
ROUGE-L-P & -0.23 & -0.21 & -0.45 & \bf -0.30 & -0.20 & -0.17 & -0.18 & -0.23 \\ 
ROUGE-L-R &  -0.20 & -0.18 & -0.33 & -0.21 & -0.19 & -0.15 & -0.09 &  -0.20 \\ 
ROUGE-L-F &  -0.24 &  -0.22 & -0.38 & -0.25 & -0.21 &  -0.18 & -0.11 & -0.20 \\ 
BERTScore-P & -0.23 & -0.21 & -0.46 &  -0.27 &  -0.22 & \bf -0.20 & -0.12 & -0.23 \\ 
BERTScore-R &  -0.22 & -0.18 & -0.31 & -0.19 & -0.22 & -0.16 & -0.05 & -0.16 \\ 
BERTScore-F & -0.24 & -0.20 & -0.39 & -0.23 & -0.22 & -0.19 & -0.08 &  -0.20\\  
BLEURT   & -0.20 & -0.20 & -0.37 &  -0.23 & -0.18 & -0.13 & -0.10 & -0.21 \\ 
BARTScore & -0.26  & -0.21 & -0.42 & -0.29 & -0.27 & -0.19 & -0.16 & -0.21 \\  \hline  
\textbf{New Metrics} &  &  &  &  &  &  &  & \\ 
MedBERT-P &  -0.23 & -0.21 & \bf -0.47 & -0.27 & -0.21 & \bf -0.20 & -0.10 & -0.23 \\ 
MedBERT-SP & -0.23 & -0.22 & \bf -0.47 &  -0.28 & -0.22 & \bf -0.20 & -0.10 & -0.23 \\ 
MedBART & -0.26 & \bf -0.23 &  -0.46 & -0.29 & -0.25 & -0.19 & -0.16 & -0.23 \\ 
ClinicalBLEURT & \bf -0.30 & -0.19 & -0.29 & -0.25 & \bf -0.31 & -0.18 & \bf -0.25 & -0.19 \\ 
MIST &  -0.07 & -0.05  & -0.12 & -0.16 & -0.09 & -0.09 & 0.02 & -0.02 \\ 
MIST-Comb1 & -0.18 & -0.15 & -0.27 & -0.20 & -0.19 & -0.15 & -0.04 & -0.13 \\ 
MIST-Comb2 & -0.18 & -0.17 & -0.30 & -0.22 & -0.18 & -0.15 & -0.06 & -0.15 \\ 
\hline  
\end{tabular}  
}
\caption{\textbf{\qafull{}}: Pearson's correlation coefficients between the automatic and manual scores on the \hpi{}, \assessment{}, \exam{}, and \results{} datasets. Unlike Tables 2-4 which present the fact-based results, here, Hallucination and Omission are measured at the key-phrase level.}
 
\label{tab:corr_qafull}
\end{table*}
%%%======
%=================================================
\section{Performance of Evaluation Metrics}
%=================================================

\begin{table*}[]
\centering
\footnotesize
\scalebox{0.95}{
\begin{tabular}{lccc|cc|c}  
\hline 
\textbf{SOTA Metrics} & $\uparrow$  \textbf{Factual P} & $\uparrow$  \textbf{Factual R} & $\uparrow$  \textbf{Factual F1} & $\downarrow$  \textbf{Hallucination} & $\downarrow$  \textbf{Omission} & $\uparrow$ \textbf{Aggregate Score} \\ 
%%%%%%%%%%%%%%%%%%%%%%%%%%%%%%%%%%%%%%%
ROUGE-1-P & 0.39 & 0.04 & 0.15 & -0.35 & -0.23 & 0.22  \\ 
ROUGE-1-R  & 0.30 & 0.64 & 0.63 & -0.20 & \underline{-0.46} & \underline{0.48} \\ 
ROUGE-1-F  & 0.38 & 0.49 & 0.55 & -0.27 & -0.43 & 0.45 \\ 
ROUGE-2-P  & 0.34 & 0.14 & 0.21 & -0.31 & -0.27 & 0.25 \\ 
ROUGE-2-R   & 0.29 & 0.51 & 0.51 & -0.23 & -0.41 & 0.41 \\ 
ROUGE-2-F    & 0.33 & 0.42 & 0.46 & -0.26 & -0.39 & 0.39 \\ 
ROUGE-L-P  & 0.38 & 0.04 & 0.15 & -0.34 & -0.23 & 0.22 \\ 
ROUGE-L-R   & 0.30 & 0.63 & 0.62 & -0.20 & -0.45 & 0.47  \\ 
ROUGE-L-F & 0.37 & 0.49 & 0.53 & -0.27 & -0.42 & 0.44\\
BERTScore-P  & 0.34 & 0.17 & 0.25 & -0.30 & -0.31 & 0.28 \\ 
BERTScore-R  & 0.28 & 0.66 & \underline{0.65} & -0.19 & \bf -0.47 & \bf 0.49 \\ 
BERTScore-F  & 0.35 & 0.48 & 0.52 & -0.26 & -0.44 & 0.44 \\ 
BLEURT & 0.35 & 0.54 & 0.56 & -0.25 & -0.44 & 0.45\\ 
BARTScore  & \bf 0.45 & 0.19 & 0.31 & \underline{-0.37} & -0.29 & 0.32  \\ 
\hline 
\textbf{New Metrics}  &  &  &  &  &   \\ 
MedBERTScore-P  & \underline{0.41} & 0.09 & 0.20 & -0.32 & -0.33 & 0.26  \\ 
MedBERTScore-SP & \underline{0.41} & 0.09 & 0.20 & -0.32 & -0.33 & 0.27 \\ 
MedBARTScore & \bf 0.45 & 0.17 & 0.29 & \bf -0.38 & -0.28 & 0.31 \\ 
ClinicalBLEURT & 0.17 & 0.16 & 0.13 & -0.08 & -0.15 & 0.12  \\ 
MIST  & 0.07 & 0.55 & 0.47 & -0.02 & -0.28 & 0.31\\ 
MIST-Comb1  & 0.25 & \bf 0.70 & \bf 0.66 & -0.15 & -0.45 & \underline{0.48} \\ 
MIST-Comb2  & 0.29 & \underline{0.68} & \bf 0.66 & -0.18 & \underline{-0.46} & \bf 0.49  \\ 
\hline  
\end{tabular}  
}
\caption{Average scores of the 21 automatic metrics across all datasets. Best results are highlighted in bold and second best are underlined.}
\label{tab:ranks}
\end{table*}

%%%======
We compute the Pearson correlation scores between the automatic metrics and the reference scores. When both manual factual scores ($F$), hallucination ($H$), and omission rates ($O$) are available, we compute an aggregate score: 
\begin{equation}
Aggregate Score = \frac{1}{4}(2F-H-O)
\end{equation}
The intuition behind this score is that both omissions (O) and hallucinations (H) are critical criteria but they need to be taken into account in the context of factual correctness (F). 

The results on the \mts{} dataset are presented in Table~\ref{tab:corr_mts}, where the ensemble metric MIST-Comb1 achieved the best correlation with manual scores on Factual F1, Factual Recall, and Omission Rate, with respective correlation values of 0.61, 0.64, and -0.71. The new MedBARTScore metric achieved the best correlation with human assessment for both Factual Precision and Hallucination Rate with 0.46 and -0.46 correlation values.   

Table~\ref{tab:corr_qatriples} presents the Pearson correlations between the automatic metrics and reference scores on the \qatriples{} dataset.
Compared with the results on the MTS dataset, ROUGE-N variants achieved high correlation scores in all categories. In particular, ROUGE-1-R and ROUGE-L-R have the best scores for Factual F1 and Factual Recall. ROUGE-1-F and ROUGE-L-F have the best scores for Factual Precision. ROUGE-1-P and ROUGE-L-P have the best correlations with the Hallucination Rate. BERTScore-R and the ensemble metric MIST-Comb2 achieved the highest correlations with manual scores for the Omission Rate. 

%% (Table~\ref{tab:corr_qafull} in Appendix B)

On the larger \qafull{} dataset, ROUGE-N results followed a similar pattern on the \hpi{}, \assessment{}, \exam{}, and \results{} subsets, as presented in Table~\ref{tab:corr_qafull}, with ROUGE-1-P, ROUGE-2-P, and ROUGE-L-P having the highest correlations with the Omission Rate in the \assessment{} dataset, and the Hallucination Rate in the \results{} dataset. This could be explained in part by the fact that the reference notes in the \qafull{} dataset have been created from initial drafts produced by summarization models which increases the likelihood of word overlap. 

The fine-tuned ClinicalBLEURT metric achieves the highest correlation scores for the Hallucination Rate in the \hpi{}, \exam{}, and \results{} datasets. The new medical metrics MedBERTScore-P and MedBERTScore-PS have the highest correlations for Hallucination and Omission Rates on the \assessment{}  and \exam{} datasets, respectively. 

Table~\ref{tab:corr_mediqarrs} presents the Pearson correlations between the automatic metrics and reference scores on the \mediqa{} dataset, where ROUGE-1-P has the highest correlation with Factual Precision and Hallucination Rate with 0.40 and -0.39. The new MIST metric has the highest correlation scores with Factual Recall and Factual F1 with 0.73 and 0.66, respectively.

Table~\ref{tab:ranks} presents the average scores of the 21 metrics across all datasets. On specific evaluation criteria, the new MedBARTScore metric performed the best on average on correlating with low Hallucinate Rate, with a correlation score of -0.38, and Factual Precision with an average correlation score of 0.45. Both MIST-Comb2 and BERTScore-R have the highest Aggregate Score of 0.49 followed by MIST-Comb1 and ROUGE-1-R. The same set of metrics has similar positive results on the \mts{}, \mediqa{}, and \qatriples{} datasets. Using the dataset-specific Aggregate Score, we observe that MIST-Comb1, MIST-Comb2, BERTScore-R, and ROUGE-1-R  perform well on Factual correctness while maintaining a stable/good performance on being indicative of lower hallucination and omission rates. These datasets are substantially different from each other: long clinical notes for \qatriples{}  (with 214 words/note), concise impression sections from radiology reports for \mediqa{}(with 18 words/summary), and different types of sections from different specialities for \mts{} (15 words/summary), which suggests that this set of metrics can be relied upon for the evaluation of clinical note generation.

%=================================================
\section{Conclusion}
%=================================================

While finding a relevant and generic evaluation metric for NLG systems remains a challenging task, our study shows that the solution to the problem is likely to be domain- and
task-specific. In particular, metrics that did well on capturing factual accuracy did not necessarily capture critical aspects in clinical note generation such as hallucinations and key medical fact omissions. Our experiments also show that language-model based metrics and metric ensembles can outperform SOTA N-gram based measures such as ROUGE when reference summaries are not biased. 
 The extensive measurements and new metrics evaluated in this paper are valuable for guiding decisions on which metrics will be most effective for researchers to use going forward in their Automated Medical Note Generation scenarios.

%Our results also show that ensemble metrics are more likely to be more stable across different datasets.

%For specific domains and tasks in Automated Medical Note Generation, there will often be one or two metrics that can do the best in covering the multiple important aspects of note quality. 

%to use in these different scenarios.

%Our experiments provide valuable data toward suggesting metrics to use for specific end tasks in Medical Note generation.

%Finally, going beyond metrics, we also find that dataset heterogeneity is a key element in building more reliable NLG evaluation methodologies. 

%=================================================
\section*{Limitations}
While our research and empirical results support specific evaluation metrics for the task of clinical note generation according to a given evaluation criteria, more results, including testing on additional datasets are needed to further validate these findings. 
Our manual annotations followed clear and structured guidelines, but could still contain some level of annotator bias and have an average Pearson inter-annotator-agreement of 0.67 (Tables~\ref{tab:iaa_qa} and~\ref{tab:iaa_rrs}).

\section*{Ethics Statement}
No protected health information will be released with the created annotations. Annotators were paid a fair hourly wage consistent with the practice of the state of hire.  

\section*{Acknowledgements}
We thank the anonymous reviewers and area chair for their valuable feedback. We also thank our annotators for their help with the manual evaluation. 

\bibliography{anthology,custom}
\bibliographystyle{acl_natbib}

\appendix

\section{Inter-Annotator Agreements (IAA)}
\label{sec:appendix}

\begin{table}[h!]
\centering
\footnotesize
\scalebox{0.85}{
\begin{tabular}{lccccc}
\hline
annotations & kappa & f1 & f1(tol=1) & f1(tol=2) & pearson\\
\hline
crit-ommissions & 0.29 & 0.48  & 0.65 &  0.75 & 0.75 \\
hallucinations & 0.46 & 0.73  & 0.87 & 0.92  & 0.97 \\
correct-facts & 0.12 & 0.13  & 0.30 &  0.40 & 0.79 \\
incorrect-facts & 0.58 & 0.73 & 0.90 & 1.00  & 0.89 \\
\hline
\end{tabular}
}
\caption{Averaged pairwise IAA for the annotation of 20 transcript-section pairs from the \qatriples{} dataset.}
\label{tab:iaa_qa}
\end{table}

\begin{table}[h!]
\centering
\footnotesize
\scalebox{0.85}{
\begin{tabular}{lccccc}
\hline
annotations & kappa & f1 & f1(tol=1) & f1(tol=2) & pearson\\
\hline
crit-ommissions & 0.26 &  0.34 & 0.66 &  0.85 & 0.81 \\
hallucinations & 0.36 &  0.76 & 0.96 &  0.98 & 0.34 \\
correct-facts & 0.07 &  0.16 & 0.60 &  0.82 & 0.76 \\
incorrect-facts & 0.06 &  0.64 & 0.79 &  0.90 & 0.07 \\
\hline
\end{tabular}
}
\caption{Averaged pairwise IAA for the annotation of 34 summary-note pairs from the \mediqa{} dataset.}
\label{tab:iaa_rrs}
\end{table}

\vspace{1cm}
\section{Finetuning-based Metric: Weight scheme}

\begin{table}[h!]
\centering
\footnotesize
\begin{tabular}{lcc}
\hline
error\_type & original weight & normalized weight\\
\hline
critical & 3 & 1  \\
non-critical & 1 & $\frac{1}{3}$  \\
spelling/grammar & $\frac{1}{4}$ & $\frac{1}{12}$  \\
\hline
\end{tabular}
\caption{Error score weights used in production for evaluating produced notes during a QA review. The normalized versions of the weights are used in our calculations so that the number of errors will not exceed over 1 per sentence unless there is more than 1 critical error.}
\label{tab:errorweights}
\end{table}

%%%======

\end{document}